\documentclass[letterpaper, 10 pt, conference]{ieeeconf}  
\usepackage{amsmath}  
\usepackage{amsfonts} 
\usepackage{amssymb}  
\usepackage{graphicx}
\usepackage{breakurl}
\usepackage{subcaption} 
\usepackage{caption} 
\usepackage{algorithm}
\usepackage[utf8]{inputenc}
\usepackage{algpseudocode}
\usepackage{booktabs}  
\usepackage{multirow}
\usepackage{makecell} 
\usepackage{quotes}

\IEEEoverridecommandlockouts                              

\title{\LARGE \bf
Optimal Scheduling of a Dual-Arm Robot for Efficient Strawberry Harvesting in Plant Factories
}

\author{Yuankai Zhu$^{\dagger 2}$, Wenwu Lu$^{\dagger 1}$, Guoqiang Ren$^{*1}$, Haiming Gao$^{1}$, Stavros Vougioukas$^{3}$ \\ Yibin Ying$^{4}$, and Chen Peng$^{*1,4}$
\thanks{*This work was supported by the Natural Science Foundation
of Zhejiang provinces, China (Grant No. LD24C130003), and in part by the National Natural Science Foundation of China (Grant No. 62303428). ($\dagger$ authors are equally contributed to this paper; corresponding author: Guoqiang Ren, Chen Peng)}
\thanks{1. Chen Peng, Wenwu Lu, Guoqiang Ren and Haiming Gao are with the ZJU-Hangzhou Global Scientific and Technological Innovation Center, Zhejiang University,
        Hangzhou, China
        {\tt\small \{chen.peng, wenwuLu, gqren, ghm\}@zju.edu.cn}}%
\thanks{2. Yuankai Zhu is with the Department of Mechanical and Aerospace Engineering, University of California, Davis,
        One Shields Ave, Davis, USA
        {\tt\small ykzhu@ucdavis.edu}}%
\thanks{3. Stavros Vougioukas is with the Department of Biological and Agricultural Engineering, University of California, Davis,
        One Shields Ave, Davis, USA
        {\tt\footnotesize svougioukas@ucdavis.edu}}%
\thanks{4. Chen Peng and Yibin Ying is with the College Of Biosystems Engineering And Food Science, Zhejiang University,
        Hangzhou, China
        {\tt\small \{chen.peng, ybying\}@zju.edu.cn}}%
}

\begin{document}

\maketitle
\thispagestyle{empty}
\pagestyle{empty}

\begin{abstract}

Plant factory cultivation is widely recognized for its ability to optimize resource use and boost crop yields. To further increase the efficiency in these environments, we propose a mixed-integer linear programming (MILP) framework that systematically schedules and coordinates dual-arm harvesting tasks, minimizing the overall harvesting makespan based on pre-mapped fruit locations. Specifically, we focus on a specialized dual-arm harvesting robot and employ pose coverage analysis of its end effector to maximize picking reachability. Additionally, we compare the performance of the dual-arm configuration with that of a single-arm vehicle, demonstrating that the dual-arm system can nearly double efficiency when fruit densities are roughly equal on both sides. Extensive simulations show a 10–20\% increase in throughput and a significant reduction in the number of stops compared to non-optimized methods. These results underscore the advantages of an optimal scheduling approach in improving the scalability and efficiency of robotic harvesting in plant factories.

\end{abstract}

\section{INTRODUCTION} \label{section: introduction}
In response to challenges posed by land policies and significant labor shortages worldwide, plant factory cultivation has emerged as a promising solution to enhance agricultural productivity\cite{hemathilake2022agricultural}. The proliferation and advancement of these cultivation models have significantly boosted the mass and continuous production of fruits and vegetables\cite{FAO}. In those environments, robotic farming equipment has become essential for managing complex and labor-intensive horticultural tasks, enhancing efficiency, and optimizing production processes\cite{ahmed2024advancing}. By integrating robotic systems within plant factories, high efficiency in crop management tasks can be achieved, particularly in labor-intensive harvesting processes\cite{ren2023mobile}. Prototype fruit-harvesting robots have been developed and deployed for common fresh fruits such as sweet peppers\cite{lenz2024hortibot}, strawberries\cite{xiong2020autonomous}, tomatoes\cite{jun2021towards}, and cucumbers\cite{jo2024suction} in many reported researches. However, the industry is still a long way from achieving an efficient and practical harvesting solution, as numerous scientific and technological challenges persist\cite{zhou2022intelligent}. Specifically, achieving efficient harvesting planning in semi-structured environments remains a significant challenge, and addressing these issues is crucial to improving robotic efficacy and adaptability in selective harvesting applications\cite{rajendran2023towards}.
\begin{figure}[tb]
    \centering
    \includegraphics[width=0.45\textwidth]{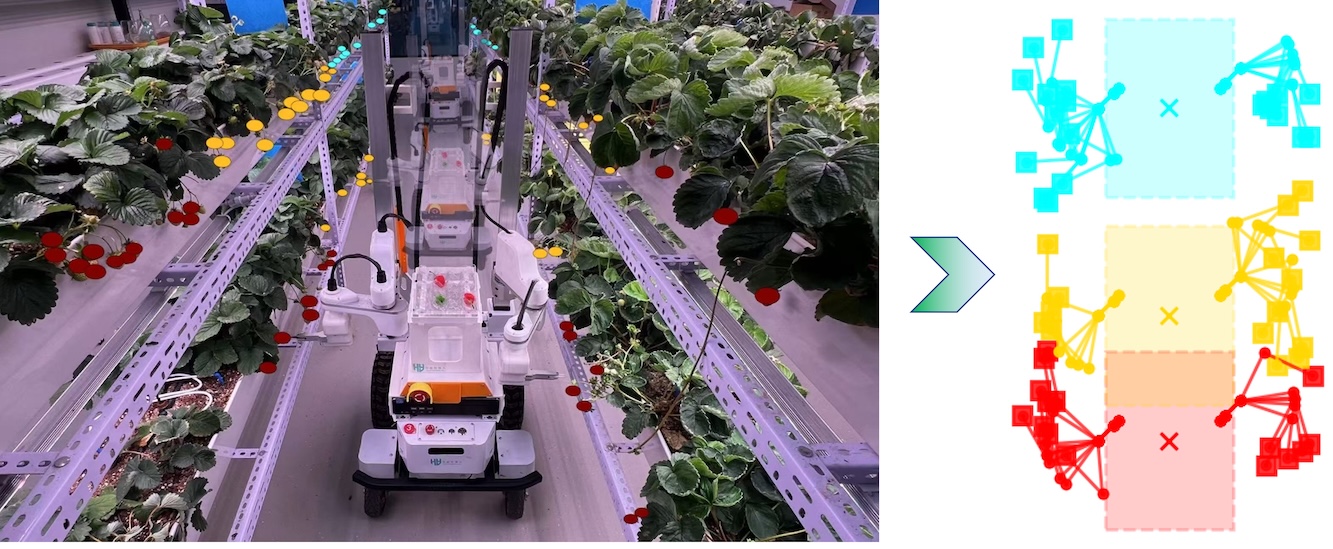}
    \caption{Stop-and-harvest strategy for strawberry picking robot with dual arms. The robot harvests some fruits on both sides at one stop, then moves to the next until all are harvested.}
    \label{fig:whole-figure}
\end{figure}

In recent years, several trends have emerged in robotic harvesting to address the challenges of efficiency in agricultural operations. One notable approach is the use of multiple robot arms harvesting simultaneously \cite{li2023multi,barnett2020work}. In this way, it becomes possible to increase the picking rate and cover larger areas within a shorter time frame, significantly boosting productivity in high-yield plant factories. Another important trend is conducting initial monitoring to generate a yield map before the harvesting process begins \cite{li2024intermittent}. This process leverages advanced perception technologies, such as Lidar and stereo cameras, to assess the quantity and location of ripe fruits within the cultivation area. Furthermore, active or multi-view vision methodologies are employed to improve the localization of fruits and address occlusion challenges \cite{magistri2024efficient, magistri2024improving, yi2024view, jose2025go}. By mapping where harvestable produce is concentrated, robots can plan their tasks more effectively, optimizing their routes and minimizing unnecessary movements. Furthermore, the data gathered during this initial inspection aids in refining perception models, which are adjusted based on local environmental conditions to enhance the accuracy of fruit detection and localization in the harvesting system \cite{fusaro2024horticultural}.

Rapid and accurate traversal inside crop rows remains challenging, in part due to the "long corridor effect," where the uniformity and repetitive structure of the row reduce visual features, making localization more difficult \cite{ren2023mobile}. To ensure reliable positioning, the robot’s speed and acceleration are typically limited, which constrains overall efficiency. While some approaches, such as Cartesian arms with visual servoing, can theoretically compensate for small positioning errors, these methods have largely remained in simulation \cite{mann2016combinatorial, pueyo2024improving}. Other arms, such as SCARA, pose greater control challenges due to limited longitudinal mobility. Thus, stopping the robot at optimized positions enables more reliable picking \cite{li2024intermittent}. Under this “stop-and-harvest” strategy, assigning fruits to each arm at each stop helps minimize total harvesting time. We propose a planner that jointly optimizes stop locations and fruit allocation to achieve this goal (see Fig.~\ref{fig:whole-figure}).

In general, the contribution of our paper can be summarized as follows:
\begin{itemize}
\item We developed a mixed-integer linear programming (MILP) model that can be solved efficiently and optimally allocates picking tasks to each robot arm, minimizing total operation time. This model is versatile and adaptable to various types of robot arms, effectively addressing the complexities of task assignment of dual-arm harvesting in a stop-and-harvest strategy for the cultivation of vertical farming. 
\item To validate the model in a real-world context, we focus on SCARA arms for strawberry harvesting. This requires optimizing the arm's installation pose to ensure maximum reachability within the specified environment. We thoroughly analyze the arm’s mechanical constraints and operational capabilities, ensuring its effectiveness in controlled agricultural settings.
\item Extensive experiments on strawberries are conducted using simulated scenarios to evaluate the performance of the proposed approach. The experimental results highlight significant improvements in harvesting efficiency, showcasing the practical applicability of our model for plant factories, like vertical farms.
\end{itemize}

The structure of this paper is: the related work is reviewed in Section \ref{section: related work}; the problem definition, the proposed MILP model and robot arms installation are introduced in Section \ref{subsection: problem definition}, \ref{subsection: mathematics modeling with milp} and \ref{subsection: optimal installation poses of robot arms}; experiments and results are listed in Section \ref{section: experiments and results}; the conclusion is in Section \ref{section: conclusion}.

\section{RELATED WORK} \label{section: related work}
Building upon these strategies, recent research has categorized harvesting methods into two main types: "pick-and-drop" and "pick-and-collect".

"Pick-and-drop" is the most common model in robot arm harvesting problems and is typically applied to non-fragile fruits like apples \cite{li2023multi}, kiwifruits \cite{barnett2020work}, citrus \cite{li2023recognition}, and melons \cite{mann2016combinatorial}. In these cases, the end effector can drop the fruit in or near the picking location into a tube or conveyor, allowing it to move over all fruit locations to collect them efficiently and continuously. This approach enables the harvesting process to be formulated as Traveling Salesman Problems (TSP), multiple TSPs (m-TSP), Vehicle Routing Problems (VRP), and their variants \cite{barnett2020work, kurtser2020planning}. Such problems are classified as in-schedule dependencies (ID) according to the taxonomy in \cite{korsah2013comprehensive}, meaning that the cost for an agent to perform a task depends on other tasks the agent is handling. 

In contrast, "pick-and-collect" strategies are essential for fragile fruit harvesting, such as strawberries \cite{xiong2020autonomous}, tomatoes \cite{li2024intermittent}, and mushrooms \cite{yang2022research}. In these cases, the end effector must return to a fixed collection point after each pick to prevent damage to the delicate produce. This makes the problem resemble task assignment problems with no dependencies (ND), where the cost for an agent to perform a task is independent of the order of other tasks assigned to the agent \cite{korsah2013comprehensive}. While standard task assignment problems can be solved in polynomial time using methods like the Hungarian algorithm\cite{kuhn1955hungarian}, the harvesting problem introduces additional complexities. Specifically, the stop positions are also the decision variables, whose number should be minimized under the constraint that the picking areas of the robot arms at all stops must cover all the assigned crops. The covering areas of the robot arm can be achieved by analyzing its reachable distribution within the operating area \cite{makhal2018reuleaux,wang2023adaptive}. Those characteristics introduce more complexity to the harvesting problem.

\section{METHODOLOGY} \label{section: methodology}
\subsection{Problem Definition} \label{subsection: problem definition}
In this work, we propose an integrated approach for minimizing dual-arm robotic harvesting completion time by utilizing "stop-and-harvest" and "pick-and-collect" methods in the plant factories. The methodology involves two key components: one is allocating fruits to each robot arm and determining the optimal stop positions for the vehicle to minimize harvesting time; the other is applying this model to a specialized dual-arm robot for harvesting, considering its kinematic and operational constraints. In the following sections, we detail the formulation of the MILP model and its practical application in a real-world setting, demonstrating how the approach enhances the harvesting efficiency.

\subsection{Mathematics Modeling with MILP}\label{subsection: mathematics modeling with milp}
\begin{figure}[tb]
    \centering
    \begin{subfigure}[b]{0.2\textwidth}
        \includegraphics[width=\textwidth]{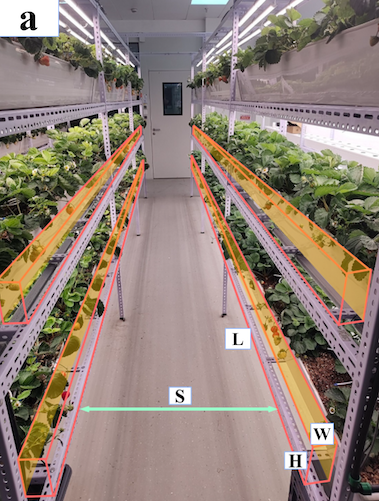}
        \label{fig:fruit_distribution}
    \end{subfigure}
    \begin{subfigure}[b]{0.2\textwidth}
        \includegraphics[width=\textwidth]{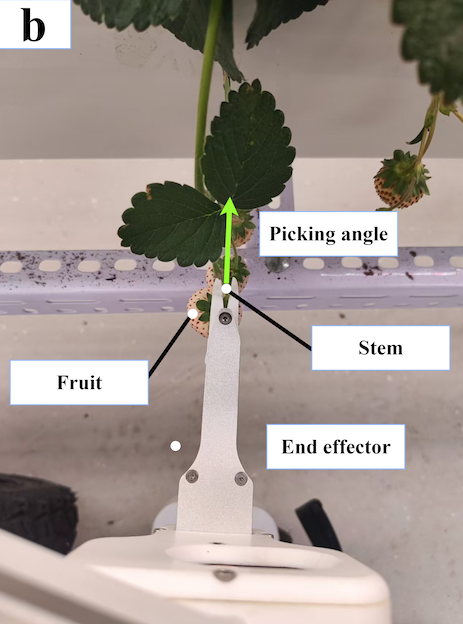}
        \label{fig:harvesting_angle}
    \end{subfigure}
    \caption{(a) Distribution of strawberries in tabletop cultivation. (b) Strawberry picking with stem-cutting.}
    \label{fig:strawberry_and_angle}
\end{figure}

\begin{figure}[tb]
    \centering
    \includegraphics[width=0.44\textwidth]{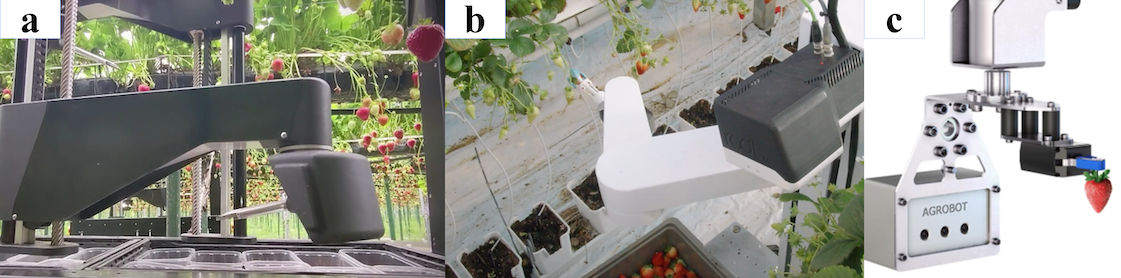}
\caption{Commercial strawberry harvesting robots with SCARA-like robot arms: 
        (a) Courtesy of Tortuga Co., Ltd., USA; 
        (b) Courtesy of Root AI Co., Ltd., USA; 
        (c) Courtesy of AgRobot Co., Ltd., USA.}
    \label{fig:other scara applications}
\end{figure}

In many industrial greenhouse cultivation systems, ripe fruits are grown in a semi-structured spatial arrangement along elevated frames (Fig.~\ref{fig:strawberry_and_angle}.a). This cultivation method inherently constrains the degrees of freedom (DoF) required for harvesting, making a 4DoF robot arm sufficient for picking fruits in 4D coordinates ($x$, $y$, $z$, and yaw $\psi$) as presented in Fig.~\ref{fig:strawberry_and_angle}.b. Previous successful implementations in similar setups have demonstrated the practicality and efficiency of using a 4DoF robot arm (see Fig.~\ref{fig:other scara applications} as examples for some commercial applications of the strawberry picking robot), such as a SCARA, for this operation \cite{xiong2020autonomous,tortuga}.

In this paper, we propose a model that is adaptable to various types of robot arms. The objective of the model is to minimize the total operation time $T$ for harvesting all fruits on both sides of plant rows. To achieve this, it is necessary to tune the number and positions of stops, and the assignment of fruits, based on the known distribution of the fruits. To choose the stop positions, we uniformly discretize the row length into N positions as candidate stop positions and select some of them. The definitions of parameters and variables in the model are provided in Table \ref{table: constants of MIP model} and Table \ref{table: variables of MIP model}.

\noindent
\textbf{Stop-and-harvest Fruit Assignment}
\begin{equation}
\min T\label{constraints: objective} \nonumber \\
\end{equation}
\text{Subject to:}\\
\begin{subequations}
\begin{align}
& T =\sum_{p=1}^N C_p + \tau\cdot \sum_{p=1}^N b_p + T_{travel} \label{constraints: objective definition}\\
& \sum_{p=1}^N a_{i, p}^L = 1, \quad \forall i \label{constraints: left fruits are only assigned once}\\
& \sum_{p=1}^N a_{j, p}^R = 1, \quad \forall j \label{constraints: right fruits are only assigned once}\\
& a_{i, p}^L \leq b_p, \quad \forall i, p \label{constraints: only harvest the left fruit at the stop if the stop is selected}\\
& a_{j, p}^R \leq b_p, \quad \forall j, p \label{constraints: only harvest the right fruit at the stop if the stop is selected}\\
& C_p^L = \sum_{i=1}^{m^L} G(x^L_i,\, y^L_i-s_p,\, z^L_i, \psi^L_i)\cdot a^L_{i, p}, \quad \forall p \label{constraints: total cost at one stop on left side} \\
& C_p^R = \sum_{i=1}^{m^R} G(x^R_i,\, y^R_i-s_p,\, z^R_i, \psi^R_j)\cdot a^R_{i, p}, \quad \forall p \label{constraints: total cost at one stop on right side} \\
& C_p = \max\{C^L_p, C^R_p\} \quad \forall p \label{constraints: harvesting time at one stop}
\end{align}
\end{subequations}


\begin{table}[tb]
\caption{MILP Model Constants Definitions}
\centering
\begin{tabular*}{\linewidth}{@{\extracolsep{\fill}}lll@{}}
\toprule
\textbf{Constants} & \textbf{Definition} & \textbf{Note} \\ 
\midrule
$s_p$ & Discretized stop positions &  \\ 
$N$ & Number of discretized stop positions & $p\in\{1...N\}$ \\ 
$\tau$ & Weights for stop numbers &  \\ 
$T_{travel}$ & Traveling time & \\
$m^L$ & Number of left fruits & $i\in\{1...m^L\}$ \\ 
$m^R$ & Number of right fruits & $j\in\{1...m^R\}$ \\ 
$G(x, y, z, \psi)$ & Cost table & \\
$x^L_i, y^L_i, z^L_i, \psi^L_i$ & The left fruit x, y, z, and yaw &  \\ 
$x^R_j, y^R_j, z^R_j, \psi^R_j$ & The right fruit x, y, z, and yaw &  \\ 
\bottomrule
\end{tabular*}
\label{table: constants of MIP model}
\end{table}

\begin{table}[tb]
\caption{MILP Model Variables Definitions}
\centering
\begin{tabular*}{\linewidth}{@{\extracolsep{\fill}}lll@{}}
\toprule
\textbf{Variables} & \textbf{Definition} & \textbf{Note} \\ 
\midrule

$b_p$ & Indicate if the stop $p$ is selected & Binary \\ 
$a^L_{i, p}$, $a^R_{j, p}$ & Indicate if fruit $i/j$ is assigned to stop $p$ & Binary \\ 
$C^L_p, C^R_p$ & Total costs at stop $p$ for left/right fruits & Continuous \\ 
$C_p$ & Total costs at stop $p$ & Continuous \\ 
\bottomrule
\end{tabular*}
\label{table: variables of MIP model}
\end{table}

This MILP model minimizes the total operation time required to pick all fruits as (\ref{constraints: objective definition}) defines. The first term is the sum of the harvesting durations of all stops. Then, due to the high accuracy required for the stop positions to ensure the fruits fall within the reachable area, the movement between stops must be slow for safety issues. Consequently, we add a weighted penalty on the number of stops, which is the re-start duration at each stop. Finally, due to the slow movement, we can regard the moving speed between stops as constant speed $v$. Therefore, the travel time along the row can be regarded as a constant $T_{travel} = \frac{L}{v}$.

The constraints (\ref{constraints: left fruits are only assigned once}) and (\ref{constraints: right fruits are only assigned once}) ensure that each fruit is assigned to be harvested only once at a stop. The constraints (\ref{constraints: only harvest the left fruit at the stop if the stop is selected}) and (\ref{constraints: only harvest the right fruit at the stop if the stop is selected}) guarantee that fruits are only harvested at a stop if that stop is selected. Additionally, (\ref{constraints: total cost at one stop on left side}) and (\ref{constraints: total cost at one stop on right side}) define the harvesting duration at one stop for the left and right sides respectively. If a candidate stop $p$ is not selected, that is, $b_p = 0$, both $C^L_p$ and $C^R_p$ are zero. (\ref{constraints: harvesting time at one stop}) defines the actual harvesting duration at each stop that is the longer duration of the left and right sides. To reduce the idle time of one side arm, the harvesting times of both sides on one-stop should be roughly equal.

To enhance the efficiency of path planning for robot arms, the configuration space cost can be pre-computed between potential fruit-picking positions and the collection position. By storing these costs in a lookup table, the system can quickly query the cost \( G(x, y, z, \psi) \) for any given position during the optimization process, avoiding the computational overhead of real-time inverse kinematics (IK) calculations.

The robot arm's reachable space is discretized into a 4D grid, where each grid point corresponds to a potential fruit-picking position. The computed costs are stored in a lookup table indexed by the discretized \( x \), \( y \), and \( z \) coordinates, as well as the picking yaw \( \psi \). Each entry in the table represents the configuration space cost calculated based on the time required for the slowest joint to reach its target position.

The motion time for each joint is computed using its maximum speed and acceleration. The overall cost for each motion is the maximum motion time across all joints, which reflects the bottleneck in the movement process. Thus, the configuration space cost is computed as:

\begin{equation}
G(x, y, z, \psi) = \max(t_1, t_2, \dots, t_n)
\end{equation}
where \( t_1, t_2, \dots, t_n \) are the motion times for all \( n \) joints.

\subsection{Optimal Installation Poses of Robot Arms} \label{subsection: optimal installation poses of robot arms}

\begin{figure}[tb]
    \centering
    \includegraphics[width=0.45\textwidth]{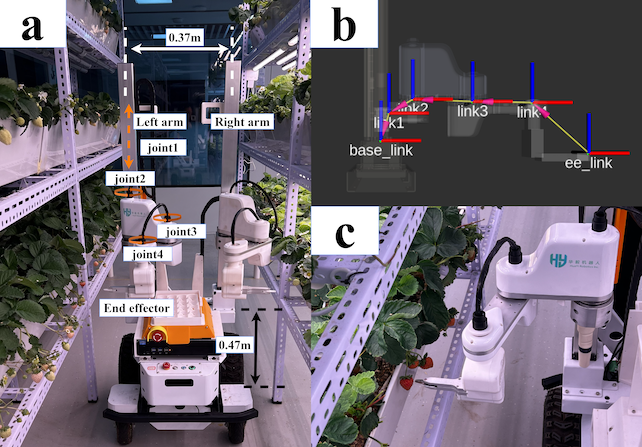}
    \caption{(a) The structure of the applied dual SCARA arm; (b) The TF tree of the applied SCARA arm when all the joint degrees are at zero; (c) The limited space for the operation of the robot arm in the greenhouse cultivation.}
    \label{fig:structure-limited-space}
\end{figure}

\begin{figure}[tb]
    \centering
    \includegraphics[width=0.4\textwidth]{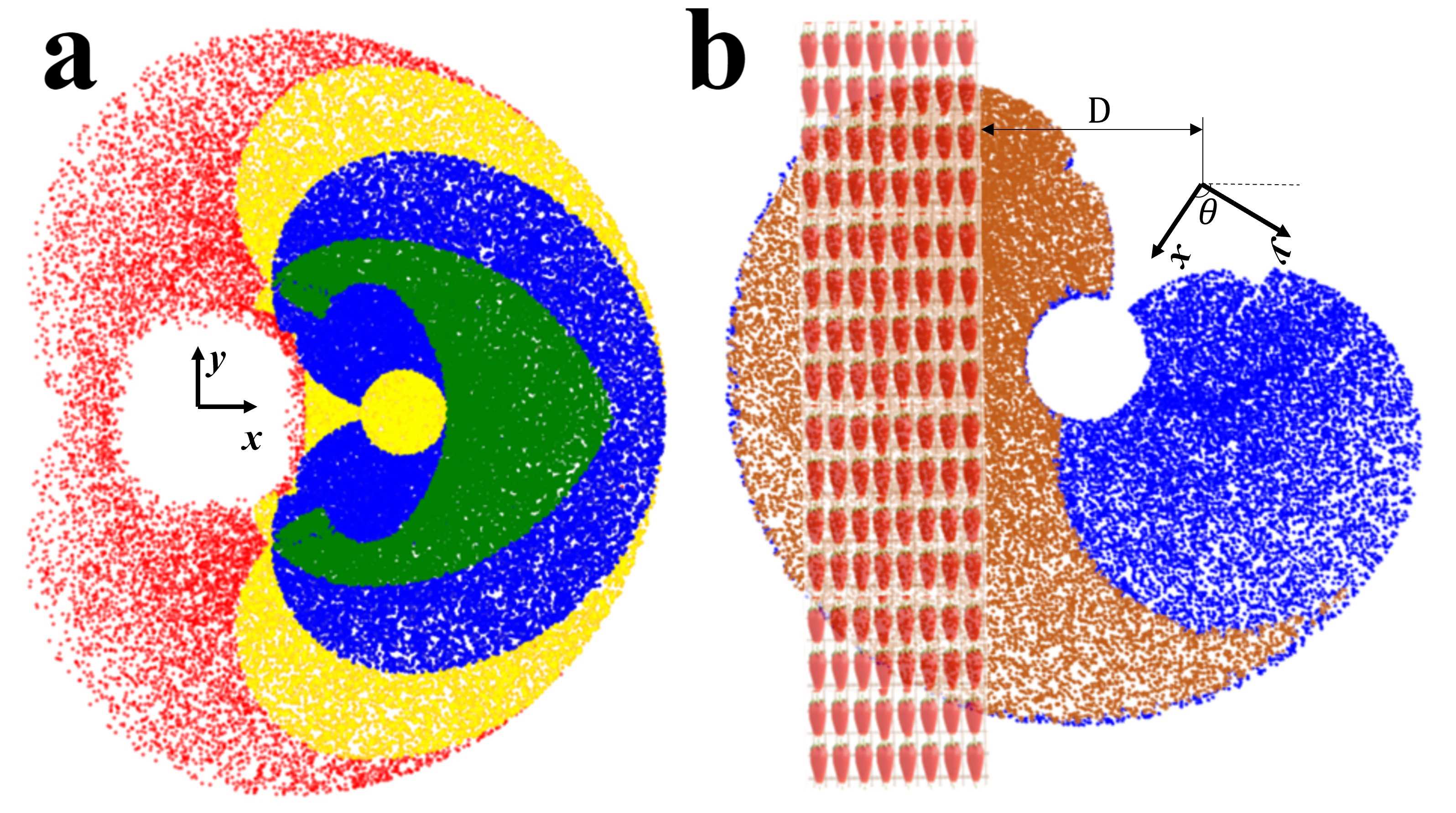}
    \caption{(a) The coverage area of different ways of picking for the SCARA robot arm: red represents the position arrival; yellow represents the end-effector can arrive in one angle among the range [-45°,45°]; blue represents the end-effector can arrive at 0°; green represents the end-effector can arrive at any angle between [-45°,45°]; (b) The coverage area where the end-effector can arrive at 0°, where orange represents the single-sided covering area.}
    \label{fig:coverage}
\end{figure}

In this study, we adopt a SCARA robot as a representative case, leveraging its compact structure, straightforward inverse kinematics, and computationally efficient motion planning shown in Fig.~\ref{fig:structure-limited-space}. These characteristics enable rapid, precise, and reliable harvesting operations. The fruit-picking type plays a significant role in determining the effective coverage area of the robot arm. Strawberries can be harvested using three main methods: stem-cutting \cite{parsa2024modular}, fruit-grasping \cite{ren2023mobile}, and bottom-suction \cite{xiong2020autonomous}. The stem-cutting or fruit-grasping method requires the end-effector to reach the stem at a specific angle, as shown in Fig.~\ref{fig:strawberry_and_angle}. In contrast, the bottom-suction method only requires the end-effector to reach the fruit at a certain position without specific angular requirements\cite{xiong2020autonomous}. These differences significantly affect the robot arm's effective coverage area as shown in Fig.~\ref{fig:coverage}, where self-collision between the applied end-effector and arms is considered. For methods requiring specific orientation (stem-cutting or fruit grasping), the robot's coverage area is limited to regions where it can achieve the necessary alignment. For the bottom-suction method, the coverage area is larger, as the end-effector only needs to reach the fruit's position, but the potential interference with the surrounding environments may limit its usage in strawberry plant factories (see Fig.~\ref{fig:structure-limited-space}.c).

Given the very compact area in which the SCARA robot operates (Fig.~\ref{fig:structure-limited-space}.c), using the full range of the arm may require adding spatial constraints to avoid collisions with the frames during high-level planning. However, by restricting the rotation angle of joint 3 (as shown in Fig.~\ref{fig:structure-limited-space}.a) to a certain range—keeping it fixed to either the left or right side—we can simplify the planning process and avoid the need for high-level algorithms \cite{rehiara2011kinematics}. This constraint eliminates the complexity of managing the arm’s multiple potential configurations in the confined space (Fig.~\ref{fig:structure-limited-space}.c). Although this limitation slightly reduces the robot arm's harvesting range, an optimal pipeline to maximize coverage within the restricted angle still ensures that the robot arm can cover as large an area as possible while simplifying the overall system.

\begin{figure}[tb]
    \centering
    \includegraphics[width=0.45\textwidth]{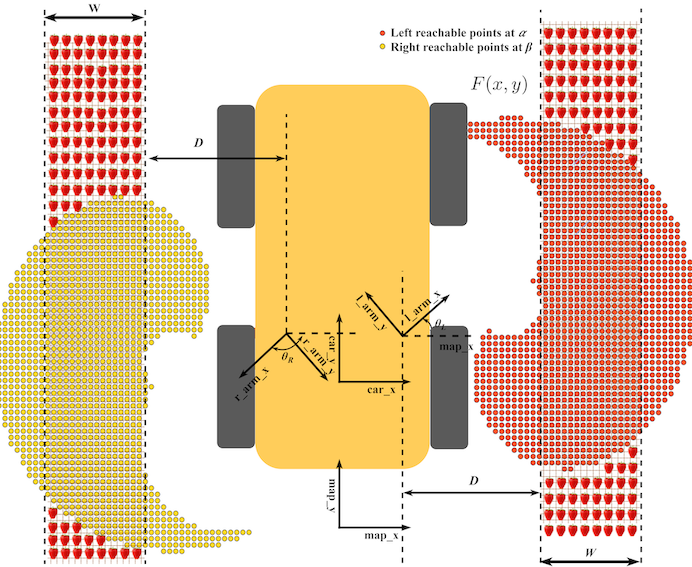}
    \caption{Reachable workspace of the SCARA robot arm in a cross-sectional plane, highlighting adjustments in the horizontal position ($D$) and rotation angles ($\theta_\mathrm{L}$ and $\theta_\mathrm{R}$) for optimal alignment with the fruit distribution.}
    \label{fig:workspace_analysis}
\end{figure}

The SCARA robot arm's manipulator workspace exhibits a cylindrical distribution. Therefore, analyzing its cross-sectional picking area suffices to determine its entire picking spatial distribution. The cross-sectional picking area of the SCARA robot arm, when intersected with the fruit-planted area (FPA), determines the actual pickable region the arm can reach. The robot arm's pose relative to the fruit-planted area influences the dimension and shape of the pick-able region. By analyzing how the arm's picking area overlaps with the fruit-planted area, we can optimize the robot arm's installation parameters, the horizontal distance ($D$) of its base along the x-axis and its rotation angles($\theta_\mathrm{L}$ and $\theta_\mathrm{R}$)  around the z-axis. This optimization maximizes the coverage area of the robot arm, ensuring it can reach the largest possible number of fruits at its installation position (Fig.~\ref{fig:workspace_analysis}).

To determine the optimal placement and orientation of the SCARA robot arm on the mobile platform, we begin by discretizing the Fruit Picking Area (FPA) into a fine grid of candidate fruit locations. We then consider a range of horizontal offsets ($D$) and rotation angles ($\theta_\mathrm{L}$ and $\theta_\mathrm{R}$) for the robot arm’s base, each discretized within predefined intervals. For each combination of $D$, $\theta_\mathrm{L}$, and $\theta_\mathrm{R}$, we solve the inverse kinematics (IK) problem to assess whether the arm can reach each sampled point in the FPA from a fixed starting pose, where the picked strawberry is placed into a container given the ND nature of the problem as mentioned above. Poses $(x,y,z,\psi)$ that yield valid IK solutions are classified as reachable. By systematically evaluating all candidate configurations, we identify the parameters ($D_{opt}$, $\theta_{\mathrm{L,opt}}$, and $\theta_{\mathrm{R,opt}}$) that maximize the number of reachable points, thus ensuring that the SCARA arm’s installation position and orientation provide the greatest possible coverage of the fruit-planted area. After deciding on the installation pose of the robot arms, the cost table of each arm can be built with IK-calculation in FPA, where the reachable pose is the value of arrival time from the start pose of the robot arm.


    

\section{EXPERIMENTS AND RESULTS} \label{section: experiments and results}
\begin{figure}[tb]
    \centering
    \includegraphics[width=0.45\textwidth]{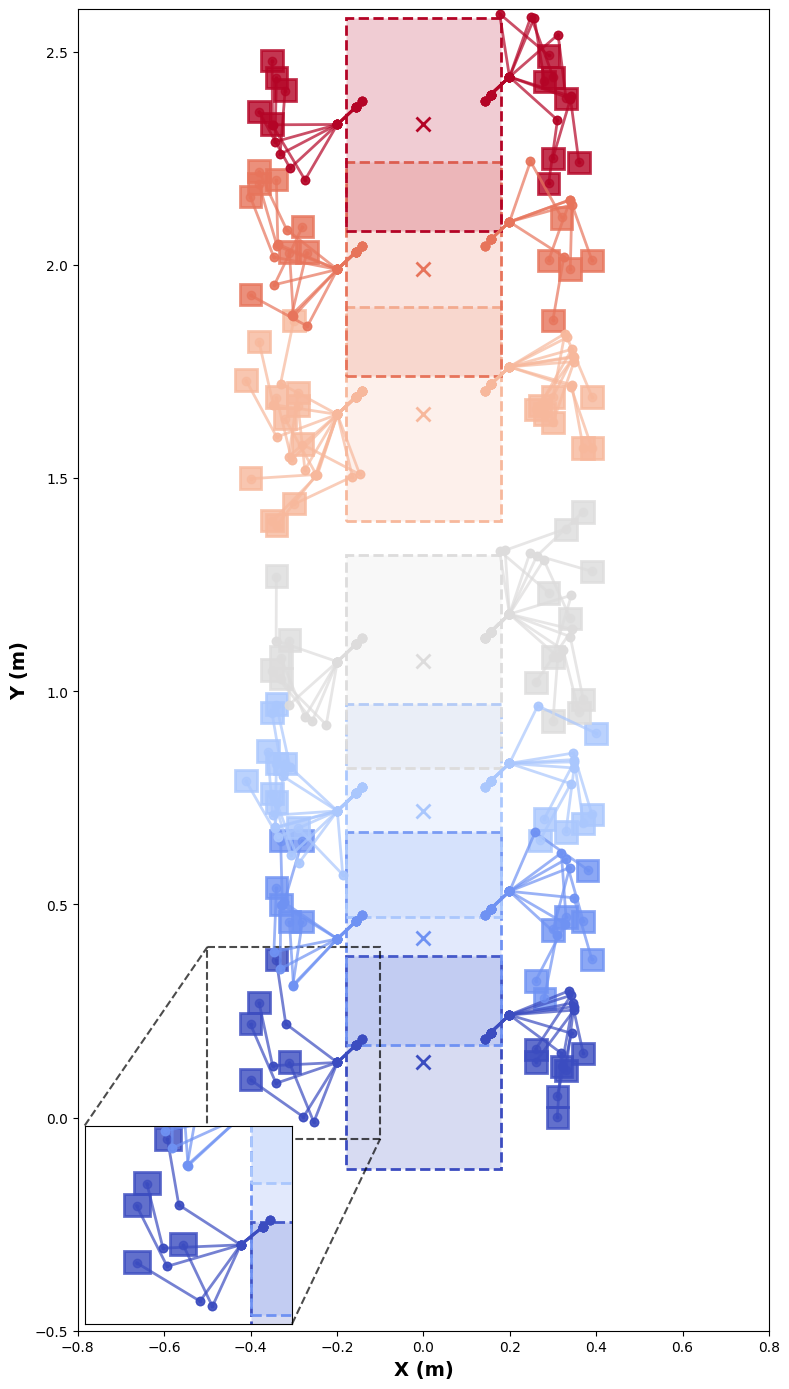}
    \caption{The output of our proposed method will output the optimal stop positions and the associated fruits relative to each stop position.}
    \label{fig:example_result}
\end{figure}

In this section, we visualize the results of the proposed method and present experiments demonstrating its significant improvement in harvesting efficiency over alternative approaches. Specifically, the model outputs the robot stop positions and the fruits assigned to each robot arm at each stop, as shown in Fig.~\ref{fig:example_result}.

\subsection{Experimental Setup}
\label{subsection: experimental setup}
The dimensions of the tested SCARA arm are shown in Fig.~\ref{fig:structure-limited-space}.a. The robot arm's base link is mounted on the robot platform at a height of 0.37 m, with the distance between the bases of the two arms being 0.47 m. When all joint values are set to zero, the relative positions of each link are shown in Fig.~\ref{fig:structure-limited-space}.b, and their origin coordinates (unit is meter) relative to the arm's base are as follows: link1 (0.02, 0.0, 0.07), link2 (0.08, 0.00, 0.10), link3 (0.23, 0.00, 0.10), link4 (0.38, 0.00, 0.10), and ee\_link (0.53, 0.00, -0.03). The speed and acceleration limits for each joint angle used in the experiment are provided in Table~\ref{table:joint_angles_and_limits}. The modeled MILP is solved with Gurobi~\cite{gurobi} and the simulations are conducted on an AMD 7950 processor with 128GB of memory.

\begin{table}[tb]
\caption{Parameters of the SCARA}
\centering
\begin{tabular*}{\columnwidth}{@{\extracolsep{\fill}}ccccc@{}}
\toprule
\makecell{\textbf{Joint} \\ \textbf{Name}} & \textbf{Type} & \makecell{\textbf{Joint} \\ \textbf{Range}} & \makecell{\textbf{Max} \\ \textbf{Speed}} & \makecell{\textbf{Max} \\ \textbf{Acceleration}}\\ 
\midrule
Joint 1 & Linear (m)       & [0, 0.5]                             & 0.1 & 0.1 \\ 
Joint 2 & Rotation (rad)   & [-1.57, 1.57]               & 0.2 & 0.2 \\ 
Joint 3 & Rotation (rad)   & [0, 2.8]  & 0.2 & 0.2 \\ 
Joint 4 & Rotation (rad)   & [-3.14, 3.14]  & 0.2 & 0.2 \\ 
\bottomrule
\end{tabular*}
\label{table:joint_angles_and_limits}
\end{table}

\subsubsection{Data Generation} \label{subsubsection: data generation}
To evaluate the proposed framework, synthetic strawberry distributions were generated in two separate rows (left and right). The left row is fixed to contain $m^L=$ 50 fruits, while the number of fruits in the right row was set to $\alpha \cdot m^L$, where $\alpha \in \{0.25, 0.4, 1.0, 2.5, 4.0\}$. Each configuration is repeated $10$ times with different random seeds to ensure statistical reliability. The spatial positions and orientations of fruits in each row are generated using a uniform random distribution within predefined bounds: row width is set as 0.7m, row length as 2m, canopy depth as 0.15m, canopy height as 0.4m given a typical setting for the greenhouse setting. We choose the discrete distance resolution as 0.01m. Therefore, there are $N=201$ candidate stop positions. The transition time from stop to move is assumed to be $\tau=5$s, and the moving speed is 0.1 m/s which indicates $T_m=20$s. 

\subsubsection{Optimal Installation Poses of Robot Arms} \label{subsubsection: optimal installation poses of robot arms}
\begin{figure}[tb] 
    \centering
    \includegraphics[width=0.5\textwidth]{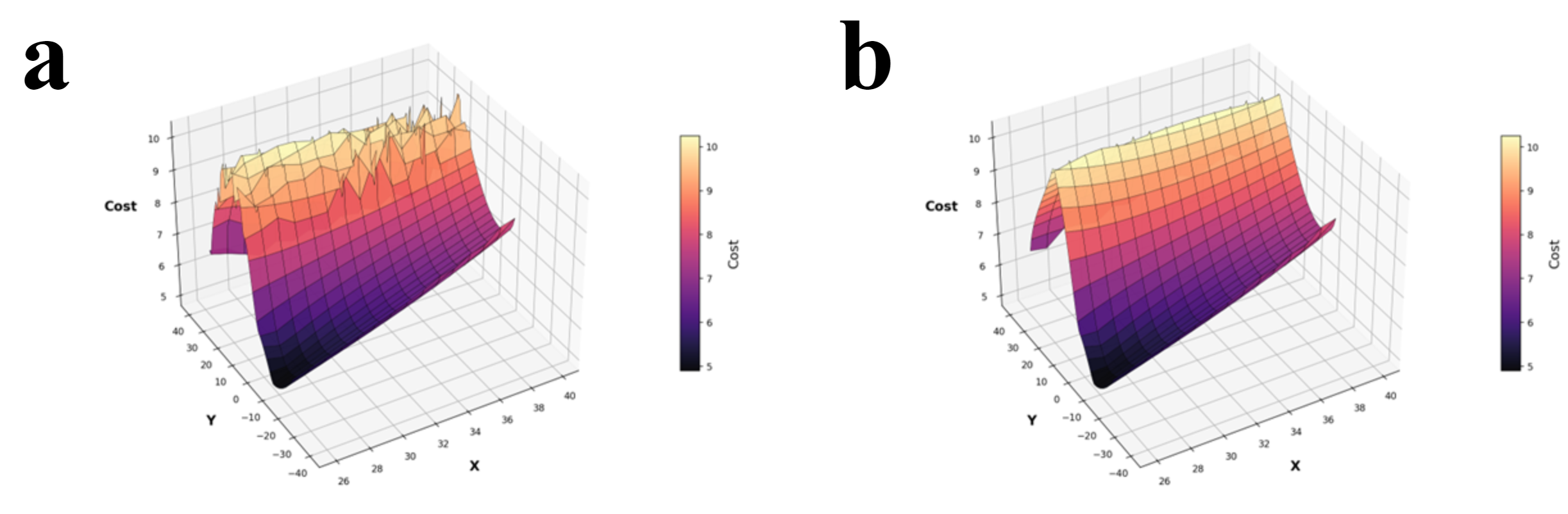}
    \caption{Cost function $G(x,y,z,\psi)$ when joint 3 is (a) unlimited or (b) limited.}
    \label{fig:cost}
\end{figure}

First, the FPA was discretized. The distance resolution was chosen as 0.01 m for both the discretized stopping positions and the $x$, $y$, and $z$ coordinates in the cost function $G$. The searching angle step for the optimal installation pose is set as $5{\circ}$. Second, the optimal installation pose of the robot arm is decided based upon the method mentioned above, the offset distance is calculated at $0.26$m and $\alpha$ is at $43.5^{\circ}$. We also visualize the cost $G$ when the joint 3 is limited or not limited (presented as Fig.~\ref{fig:cost}), from which we can see that the cost function will not be smooth, which may affect the solving time of MILP. This also supports us to apply "single-sided" settings as mentioned above. 

\subsubsection{Evaluation Metrics}\label{subsubsection: evaluation metrics}
We report three major metrics:
\begin{itemize}
    \item \textbf{Number of Stops}: The total number of stopping positions required for the harvester to harvest in each method.
    \item \textbf{Throughput}: The ratio of the total harvested fruits to the total operational time as shown in Eq.\ref{eq:throughput}. A higher throughput implies more fruits harvested per unit of time.
    \begin{equation}
        \textit{Throughputs} = \frac{m_L+m_R}{T}
        \label{eq:throughput}
    \end{equation}
    \item \textbf{Runtime}: 
    Although the fruits are pre-mapped, the running time of the planning algorithm is critical for two main reasons. First, online updates may be needed to incorporate newly detected strawberries. Second, since the robot can start or resume from any point along a row, dynamic replanning is essential. In actual deployment, the planner continuously updates based on the latest fruit detections to plan for the next two meters of harvest.
\end{itemize}

\subsection{Comparison of Harvesting Strategies}
\label{subsection: comparison of harvesting strategies}

Three harvesting strategies are evaluated and compared in the designed experiments, which can be illustrated as follows:

\begin{enumerate}
    \item \textbf{Non-optimized harvesting (FOV)}: \\
    This strategy estimates the stereo camera's coverage based on its field of view (FOV) which is smaller than the reachability area. Assuming that the pre-mapped fruit data is agnostic, the robot moves exactly a distance equal to the camera's FOV along the row after collecting all fruits in the current FOV. 
    
    \item \textbf{Optimized Single-arm harvesting (Serial)}: \\
    In this approach, the left and right rows are treated as independent tasks. The harvester completes one side of a row entirely before switching to the other, effectively processing each row in sequence. The Non-optimized solution is served as a warm start for the solver of MIP to accelerate its solving progress.
    
    \item \textbf{Optimized dual-arm harvesting (MILP)}: \\
    Our novel approach applies an MILP model to jointly determine optimal stop positions and fruit assignments, thus minimizing the total harvesting duration.
\end{enumerate}

In each method, all three evaluation metrics are recorded. This comprehensive evaluation allows us to statistically compare the efficiency and effectiveness of each strategy.

\subsection{Results and Analysis}
\label{subsection: results and analysis}

\subsubsection{Quantitative Comparison}\label{subsubsection: quantitative comparison}
\begin{figure}[tb]
    \centering
    \begin{subfigure}[b]{0.42\textwidth}
        \centering
        \includegraphics[width=\textwidth]{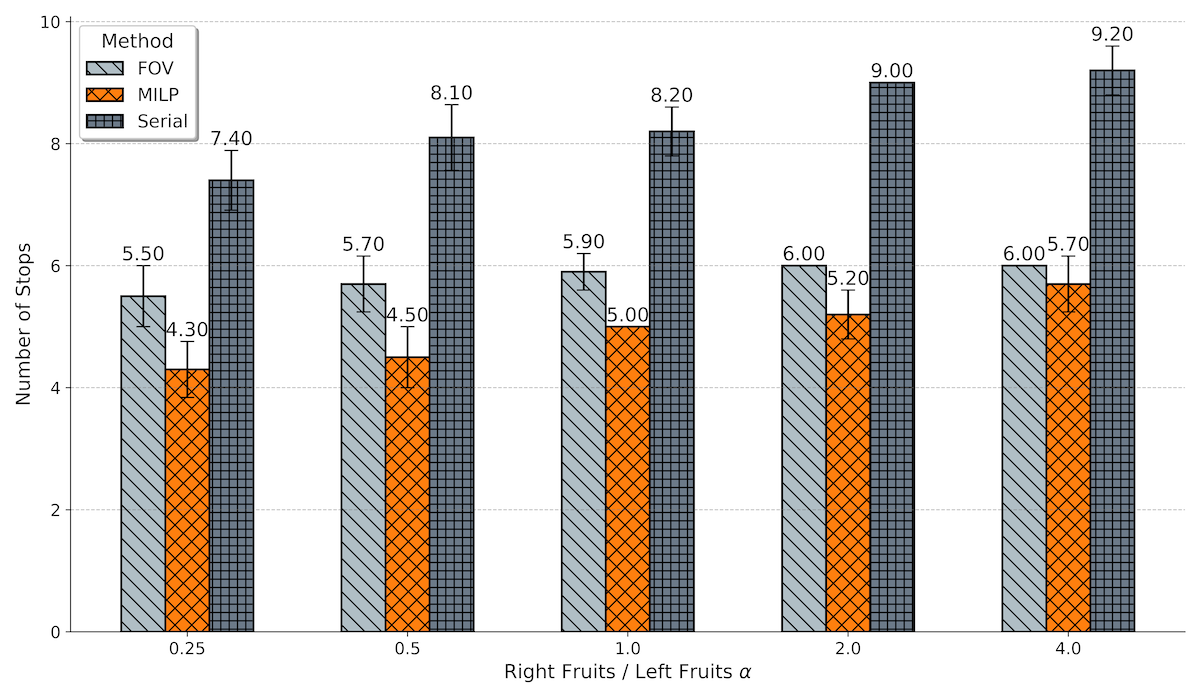}
        \caption{Number of Stops}
        \label{fig:stop_number_compare}
    \end{subfigure}
    \hfill
    \centering
    \begin{subfigure}[b]{0.42\textwidth}
        \centering
        \includegraphics[width=\textwidth]{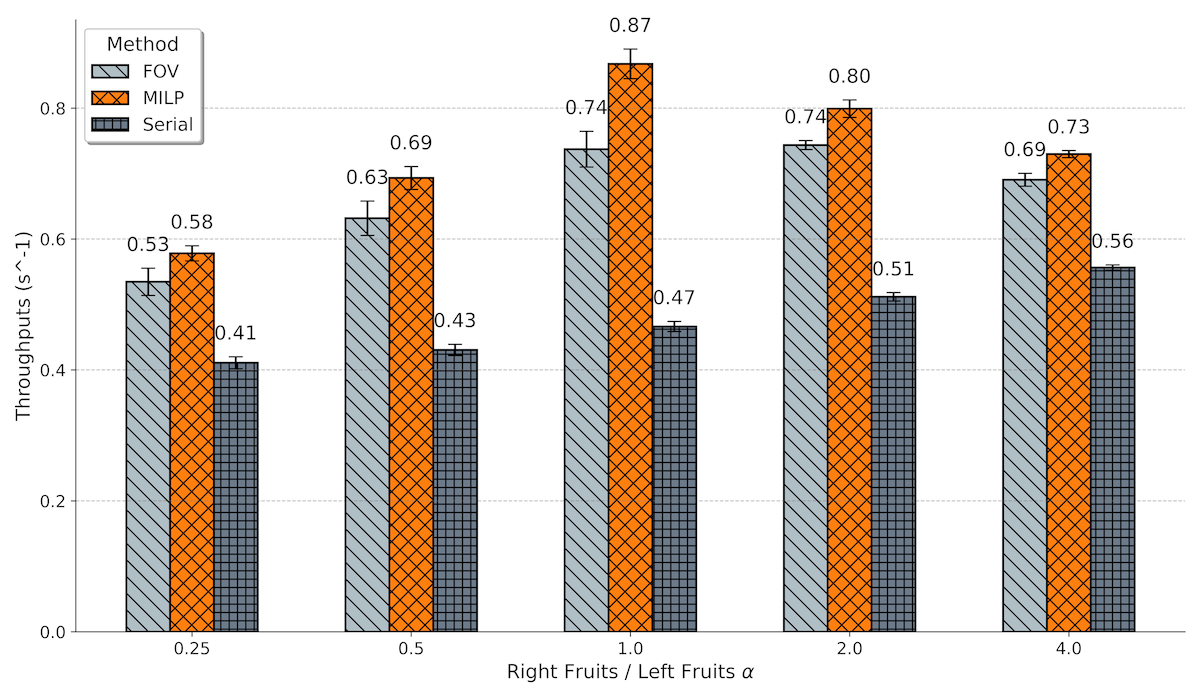}
        \caption{Throughputs.}
        \label{fig:throughput_compare}
    \end{subfigure}
    \hfill
    \centering
    \begin{subfigure}[b]{0.42\textwidth}
        \centering
        \includegraphics[width=\textwidth]{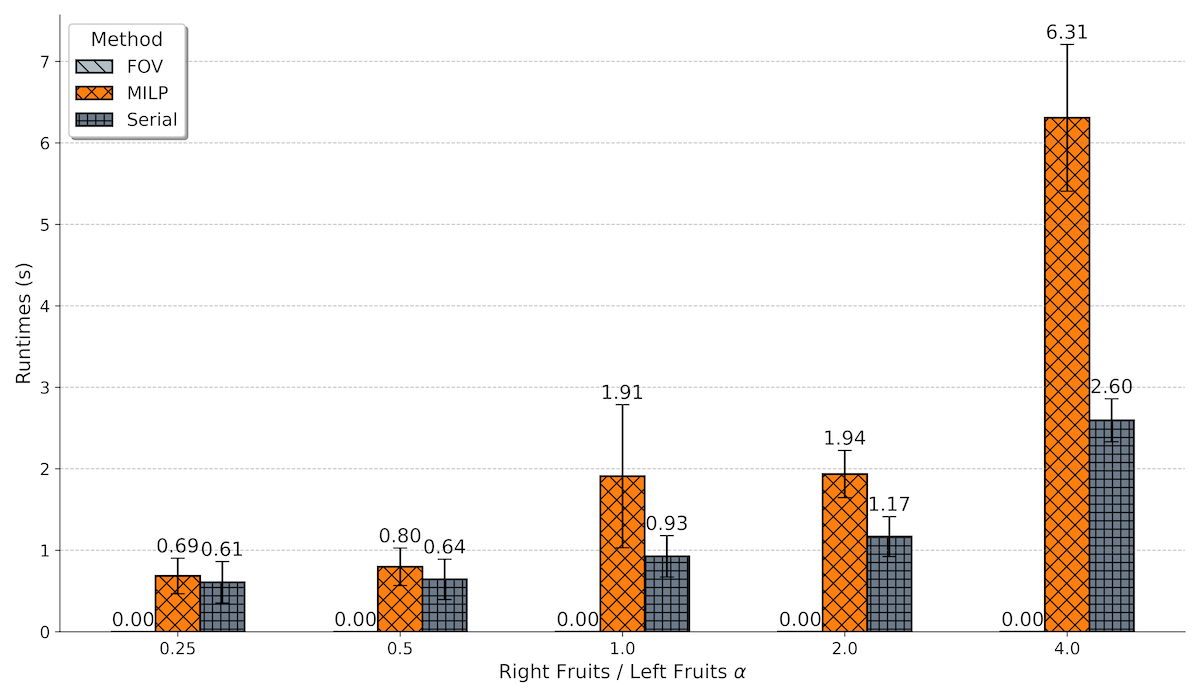}
        \caption{Runtimes.}
        \label{fig:runtime_compare}
    \end{subfigure}
    \hfill
    \caption{Performance comparison of different harvesting strategies with respect to (a) Number of Stops, (b) Throughputs, and (c) Runtimes, across various $\alpha$ values.}
    \label{fig:boxplot_compare}
\end{figure}

Fig.~\ref{fig:boxplot_compare} illustrates the distributions of the performance metrics under varying right-to-left ratios ($\alpha$). Each box plot aggregates the results from $10$ independent runs per ratio.

In terms of \emph{number of stops} (see Fig.~\ref{fig:stop_number_compare}), the proposed MILP method consistently requires the fewest stops among all tested strategies. As the total number of fruits increases (i.e., with an increase in $\alpha$), the number of stops rises accordingly.

Regarding \emph{throughput} (see Fig.~\ref{fig:throughput_compare}), the MILP method achieves superior performance, especially when $\alpha = 1.0$. This is because a balanced fruit distribution on both sides facilitates equalization of the harvesting durations on each side ($C^L_p$ and $C^R_p$) at a single stop. As a result, the ideal time for an arm to wait for the other arm is much less compared to the other cases.

Concerning \emph{runtime}, the non-optimized method is the fastest among the three approaches. Although the proposed MILP method has the longest runtime, it remains within approximately 7s. The runtime increases with the total number of fruits, as a higher fruit count leads to more decision variables and parameters, thus increasing model complexity.




\subsubsection{Discussion}\label{subsubsection: discussion}
Overall, the proposed MILP method demonstrates robust performance improvements in both the number of stops and throughput enhancement within a reasonable runtime, especially in the case $\alpha$ is near $1.0$. However, when $\alpha$ is far from $1.0$, the improvement of the proposed method is not significant. In addition, longer row lengths and larger fruit numbers would increase the complexity of solving the problem. 


\section{CONCLUSION AND FUTURE WORK}
\label{section: conclusion}

In this paper, we introduced a novel mixed-integer linear programming (MILP) formulation for the multi-row strawberry harvesting task. Our formulation jointly optimizes stop positions and picking assignments within a unified framework to minimize the total operation time. Extensive experiments, conducted under varying fruit distribution ratios and multiple random seeds, empirically demonstrate that our proposed method outperforms both non-optimized and optimized single-arm harvesting strategies in terms of the number of stops and throughput. Future work will focus on deploying and testing the proposed algorithm on a real robot system integrated with a pre-mapping algorithm, to validate its performance under practical conditions. In addition, we plan to extend our formulation by incorporating additional degrees of freedom for the robot arm and eliminating the ideal waiting time between the dual arms. This enhancement will enable both arms to work simultaneously on the same side or opposite sides, further improving the harvesting efficiency in real-world scenarios. Furthermore, we are building the prototype dual-arm harvesting robot to validate our proposed work in the real plant factories.

\addtolength{\textheight}{-0cm}   





\section*{ACKNOWLEDGMENT}
This work is supported by the Natural Science Foundation of Zhejiang provinces, China (Grant No. LD24C130003)

\bibliographystyle{IEEEtran}
\bibliography{IEEEabrv, reference}

\end{document}